\documentclass[10pt,twocolumn,letterpaper]{article}

\usepackage[pagenumbers]{cvpr} 

\usepackage[dvipsnames]{xcolor}

\usepackage{algorithm}
\usepackage[noend]{algpseudocode}
\usepackage[misc]{ifsym}
\definecolor{cvprblue}{rgb}{0.21,0.49,0.74}
\usepackage[pagebackref,breaklinks,colorlinks,citecolor=cvprblue]{hyperref}

\def\paperID{11193} 
\def\confName{CVPR}
\def\confYear{2024}

\def\METHODNAME{\textsc{FreSCo}}

\title{\METHODNAME: Spatial-Temporal Correspondence for Zero-Shot Video Translation\vspace{-2mm}}

\author{Shuai Yang$^1$\footnotemark \hspace{12pt} Yifan Zhou$^2$ \hspace{12pt}  Ziwei Liu$^2$   \hspace{12pt} Chen Change Loy$^{2~\textrm{\Letter}}$\\
\normalsize{$^1 $Wangxuan Institute of Computer Technology, Peking University \hspace{12pt}
$^2 $S-Lab, Nanyang Technological University}\\
{\tt\small williamyang@pku.edu.cn\hspace{12pt}\{yifan006,  ziwei.liu, ccloy\}@ntu.edu.sg}\vspace{-1mm}
}

\newcommand{\CR}{\textcolor{black}}

\begin{document}

\twocolumn[{%
\renewcommand\twocolumn[1][]{#1}%
\maketitle
\vspace{-2em}
\begin{center}
\centering
\includegraphics[width=0.95\linewidth]{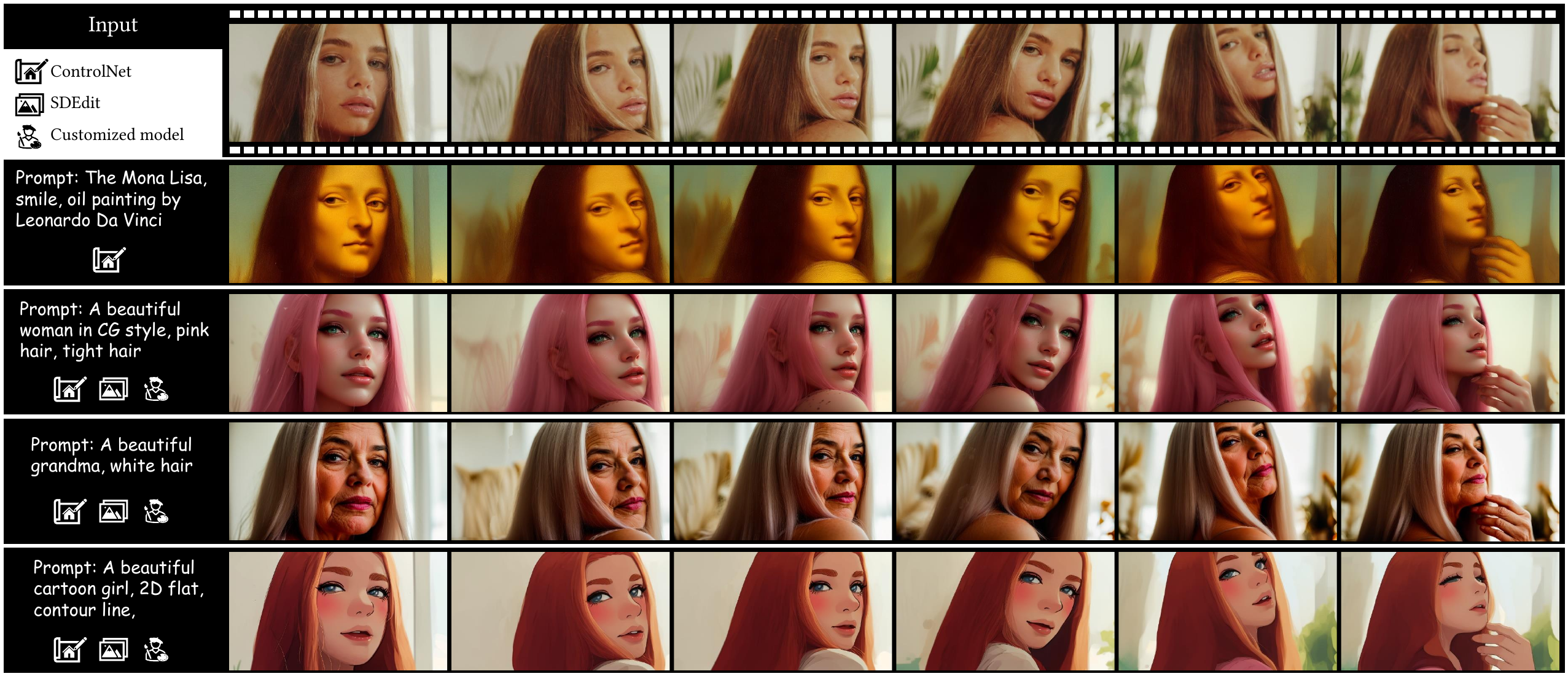}\vspace{-0.8em}
\captionof{figure}{Our framework enables high-quality and coherent video translation based on pre-trained image diffusion model. Given an input video, our method re-renders it based on a target text prompt, while preserving its semantic content and motion. Our zero-shot framework is compatible with various assistive techniques like ControlNet, SDEdit and LoRA, enabling more flexible and customized translation.}
\label{fig:teaser}
\end{center}%
}]

\footnotetext{* Work done when Shuai Yang was RAP at S-Lab, NTU.}

\begin{abstract}

The remarkable efficacy of text-to-image diffusion models has motivated extensive exploration of their potential application in video domains.
Zero-shot methods seek to extend image diffusion models to videos without necessitating model training.
Recent methods mainly focus on incorporating inter-frame correspondence into attention mechanisms. However, the soft constraint imposed on determining where to attend to valid features can sometimes be insufficient, resulting in temporal inconsistency.
In this paper, we introduce \textbf{FRESCO}, intra-frame correspondence alongside inter-frame correspondence to establish a more robust spatial-temporal constraint. This enhancement ensures a more consistent transformation of semantically similar content across frames. Beyond mere attention guidance, our approach involves an explicit update of features to achieve high spatial-temporal consistency with the input video, significantly improving the visual coherence of the resulting translated videos.
Extensive experiments demonstrate the effectiveness of our proposed framework in producing high-quality, coherent videos, marking a notable improvement over existing zero-shot methods.

\end{abstract}

\section{Introduction}
\label{sec:intro}

In today's digital age, short videos have emerged as a dominant form of entertainment. The editing and artistic rendering of these videos hold considerable practical importance. Recent advancements in diffusion models~\cite{rombach2022high,ramesh2022hierarchical,saharia2022photorealistic} have revolutionized image editing by enabling users to manipulate images conveniently through natural language prompts. Despite these strides in the image domain, video manipulation continues to pose unique challenges, especially in ensuring natural motion with temporal consistency.


Temporal-coherent motions can be learned by training video models on extensive video datasets~\cite{ho2022imagen,singer2022make,esser2023structure} or finetuning refactored image models on a single video~\cite{wu2023tune,shin2023edit,liu2023video}, which is however neither cost-effective nor convenient for ordinary users.
Alternatively, zero-shot methods~\cite{qi2023fatezero,wang2023zero,ceylan2023pix2video,geyer2023tokenflow,cong2023flatten,khachatryan2023text2video,yang2023rerender} offer an efficient avenue for video manipulation by altering the inference process of image models with extra temporal consistency constraints.
Besides efficiency, zero-shot methods possess the advantages of high compatibility with various assistive techniques designed for image models, \eg, ControlNet~\cite{zhang2023adding} and LoRA~\cite{hulora}, enabling more flexible manipulation.

\begin{figure}[t]
\centering
\includegraphics[width=0.98\linewidth]{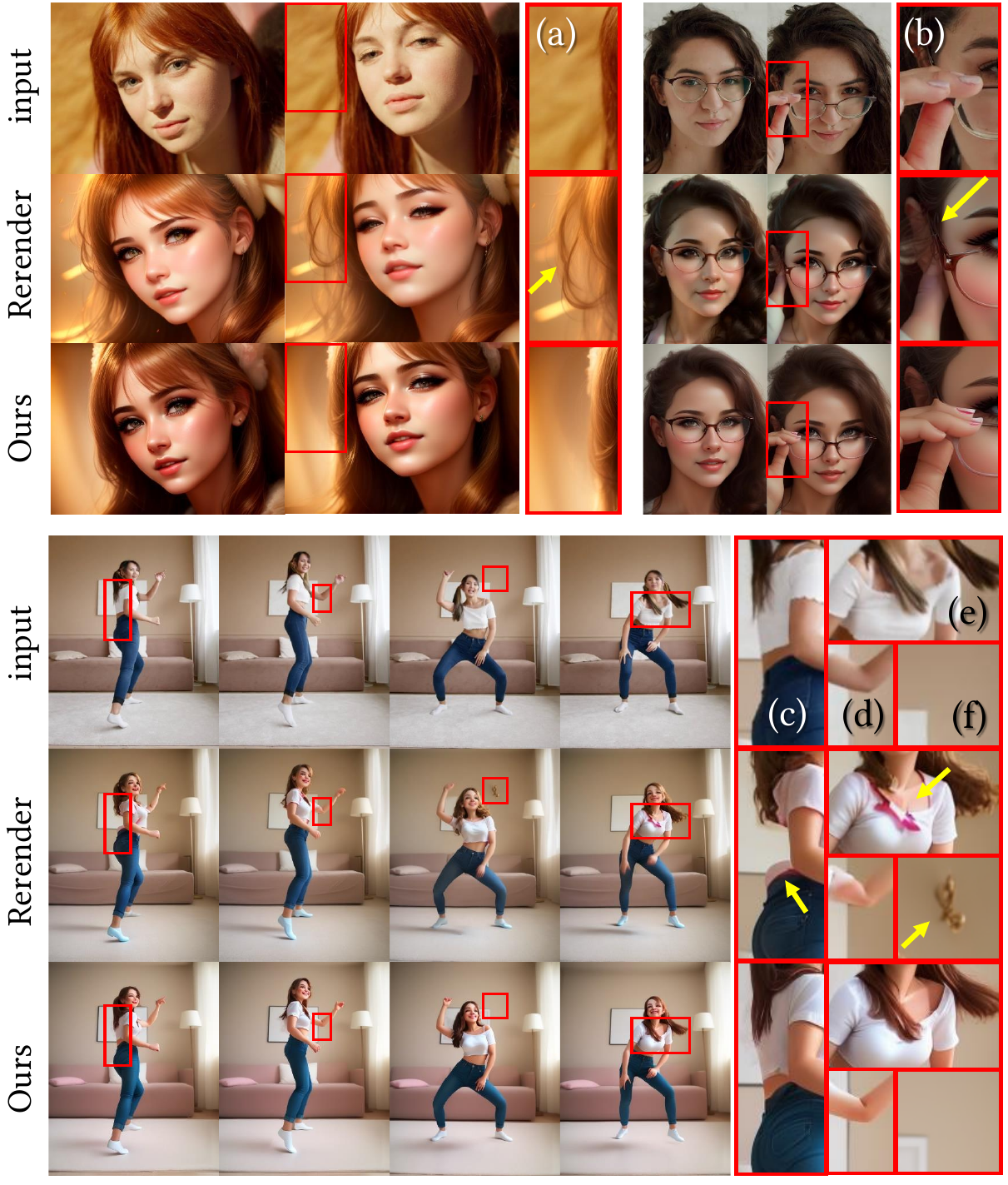}\vspace{-2mm}
\caption{Real video to CG video translation. Methods~\cite{yang2023rerender} relying on optical flow alone suffer (a)(f) inconsistent or (c)(d)(e) missing optical flow guidance and (b) error accumulation. By introducing \METHODNAME, our method addresses these challenges well.}\vspace{-5mm}
\label{fig:challenge}
\end{figure}

Existing zero-shot methods predominantly concentrate on refining attention mechanisms. These techniques often substitute self-attentions with cross-frame attentions~\cite{wu2023tune, khachatryan2023text2video}, aggregating features across multiple frames. However, this approach ensures only a coarse-level global style consistency.
To achieve more refined temporal consistency, approaches like Rerender-A-Video~\cite{yang2023rerender} and FLATTEN~\cite{cong2023flatten} assume that the generated video maintains the same inter-frame correspondence as the original. They incorporate the optical flow from the original video to guide the feature fusion process. While this strategy shows promise, three issues remain unresolved.
\textbf{1) Inconsistency.}
Changes in optical flow during manipulation may result in inconsistent guidance, leading to issues such as parts of the foreground appearing in stationary background areas without proper foreground movement (Figs.~\ref{fig:challenge}(a)(f)).
\textbf{2) Undercoverage.}
In areas where occlusion or rapid motion hinders accurate optical flow estimation, the resulting constraints are insufficient, leading to distortions as illustrated in Figs.~\ref{fig:challenge}(c)-(e).
\textbf{3) Inaccuracy.} The sequential frame-by-frame generation is restricted to local optimization, leading to the accumulation of errors over time (missing fingers in Fig.~\ref{fig:challenge}(b) due to no reference fingers in previous frames).


To address the above critical issues, we present FRamE Spatial-temporal COrrespondence (\textbf{\METHODNAME}). While previous methods primarily focus on constraining \textit{inter-frame temporal correspondence}, we believe that preserving \textit{intra-frame spatial correspondence} is equally crucial.
Our approach ensures that semantically similar content is manipulated cohesively, maintaining its similarity post-translation.
%
This strategy effectively addresses the first two challenges:
it prevents the foreground from being erroneously translated into the background, and it enhances the consistency of the optical flow.
For regions where optical flow is not available, the spatial correspondence within the original frame can serve as a regulatory mechanism, as illustrated in Fig.~\ref{fig:challenge}.

In our approach, \METHODNAME~is introduced to two levels: attention and feature.
At the attention level, we introduce \METHODNAME-guided attention. It builds upon the optical flow guidance from~\cite{cong2023flatten} and enriches the attention mechanism by integrating the self-similarity of the input frame. It allows for the effective use of both inter-frame and intra-frame cues from the input video, strategically directing the focus to valid features in a more constrained manner.
At the feature level, we present \METHODNAME-aware feature optimization. This goes beyond merely influencing feature attention; it involves an explicit update of the semantically meaningful features in the U-Net decoder layers. This is achieved through gradient descent to align closely with the high spatial-temporal consistency of the input video. The synergy of these two enhancements leads to a notable uplift in performance, as depicted in Fig.~\ref{fig:teaser}.
To overcome the final challenge, we employ a multi-frame processing strategy. Frames within a batch are processed collectively, allowing them to guide each other, while anchor frames are shared across batches to ensure inter-batch consistency. For long video translation, we use a heuristic approach for keyframe selection and employ interpolation for non-keyframe frames.
Our main contributions are:
\begin{itemize}
    \item A novel zero-shot diffusion framework guided by frame spatial-temporal correspondence for coherent and flexible video translation.
    \item Combine \METHODNAME-guided feature attention and optimization as a robust intra-and inter-frame constraint with better consistency and coverage than optical flow alone.
    \item Long video translation by jointly processing batched frames with inter-batch consistency.
\end{itemize}

\section{Related Work}
\label{sec:related_work}
\vspace{-2mm}

\noindent
\textbf{Image diffusion models.}
Recent years have witnessed the explosive growth of image diffusion models for text-guided image generation and editing. Diffusion models synthesize images through an iterative denoising process~\cite{ho2020denoising}.
DALLE-2~\cite{ramesh2022hierarchical} leverages CLIP~\cite{radford2021learning} to align text and images for text-to-image generation.
Imagen~\cite{saharia2022photorealistic} cascades diffusion models for high-resolution generation, where class-free guidance~\cite{nichol2022glide} is used to improve text conditioning.
Stable Diffusion builds upon
latent diffusion model~\cite{rombach2022high} to denoise at a compact latent space to further reduce complexity.

Text-to-image models have spawned a series of image manipulation models~\cite{hertz2022prompt,brooks2023instructpix2pix}.
Prompt2Prompt~\cite{hertz2022prompt} introduces cross-attention control to keep image layout. 
To edit real images, DDIM inversion~\cite{songdenoising} and Null-Text Inversion~\cite{mokady2023null} are proposed to embed real images into the noisy latent feature for editing with attention control~\cite{parmar2023zero,tumanyan2023plug,cao2023masactrl}.

Besides text conditioning, various flexible conditions are introduced.~SDEdit~\cite{meng2021sdedit} introduces image guidance during generation.
Object appearances and styles can be customized by finetuning text embeddings~\cite{gal2022image}, model weights~\cite{ruiz2023dreambooth,kumari2022customdiffusion,han2023svdiff,hulora} or encoders~\cite{wei2023elite,xu2023prompt,ye2023ip, gong2023talecrafter, gal2023encoder}. ControlNet~\cite{zhang2023adding} introduces a control path to provide structure or layout information for fine-grained generation. Our zero-shot framework does not alter the pre-trained model and, thus is compatible with these conditions for flexible control and customization as shown in Fig.~\ref{fig:teaser}.


\noindent
\textbf{Zero-shot text-guided video editing.}
While large video diffusion models trained or fine-tuned on videos have been studied~\cite{ho2022imagen,singer2022make,he2022latent,zhou2022magicvideo,luo2023videofusion,esser2023structure,blattmann2023align,guo2023animatediff,wu2023tune,shin2023edit,feng2023ccedit,ge2023pyoco,he2022latent,wang2023lavie}, this paper focuses on lightweight and highly compatible zero-shot methods. 
Zero-shot methods can be divided into inversion-based and inversion-free methods.

\begin{figure*}[t]
\centering
\includegraphics[width=\linewidth]{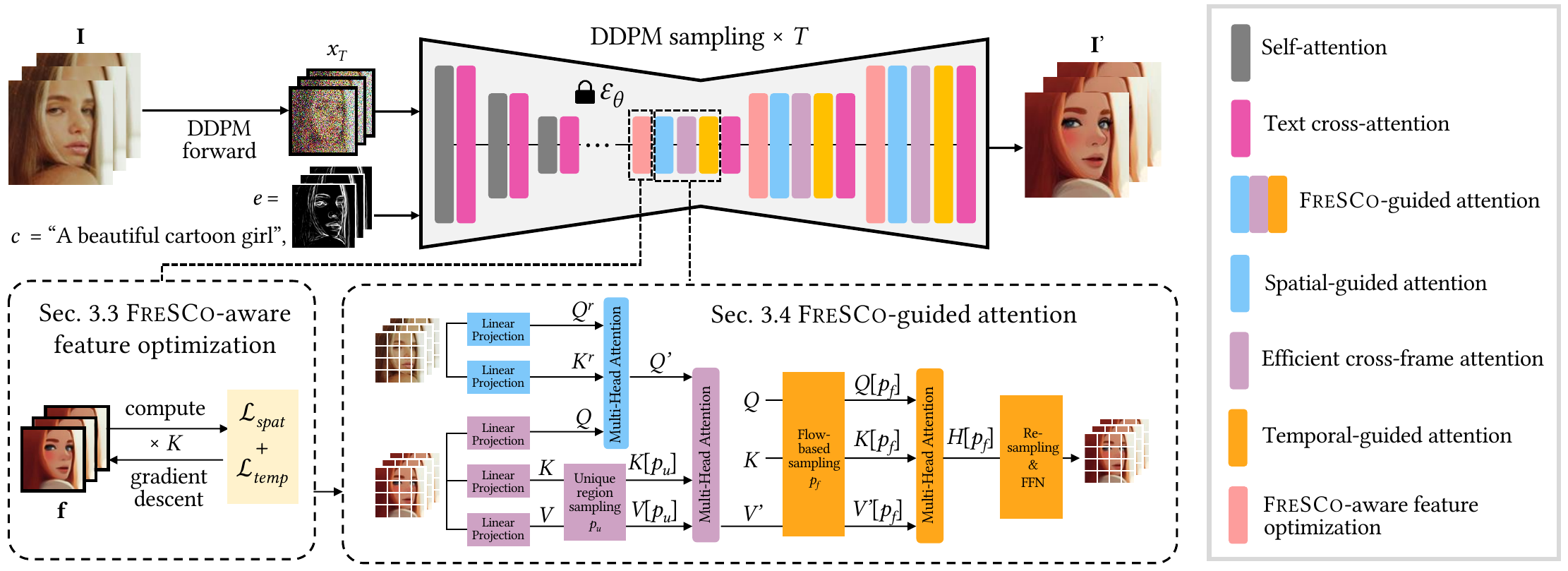}\vspace{-2mm}
\caption{Framework of our zero-shot video translation guided by FRamE Spatial-temporal COrrespondence (\METHODNAME). A \METHODNAME-aware optimization is applied to the U-Net features to strengthen their temporal and spatial coherence with the input frames. We integrate \METHODNAME~into self-attention layers, resulting in spatial-guided attention to keep spatial correspondence with the input frames, efficient cross-frame attention and temporal-guided attention to keep rough and fine temporal correspondence with the input frames, respectively.}\vspace{-3mm}
\label{fig:pipeline}
\end{figure*}

Inversion-based methods~\cite{qi2023fatezero,jeong2023ground} apply DDIM inversion to the video and record the attention features for attention control during editing.
FateZero~\cite{qi2023fatezero} detects and preserves the unedited region and uses cross-frame attention to enforce global appearance coherence.
To explicitly leverage inter-frame correspondence, Pix2Video~\cite{ceylan2023pix2video} and TokenFlow~\cite{geyer2023tokenflow} match or blend features from the previous edited frames. FLATTEN~\cite{cong2023flatten} introduces optical flows to the attention mechanism for fine-grained temporal consistency.

Inversion-free methods mainly use ControlNet for  translation.~Text2Video-Zero~\cite{khachatryan2023text2video} simulates motions by moving noises.~ControlVideo~\cite{zhang2023controlvideo} extends ControlNet to videos with cross-frame attention and inter-frame smoothing.~VideoControlNet~\cite{hu2023videocontrolnet} and Rerender-A-Video~\cite{yang2023rerender} warps and fuses the previous edited frames with optical flow to improve temporal consistency.
Compared to inversion-based methods, inversion-free methods allow for more flexible conditioning and higher compatibility with the customized models, enabling users to conveniently control the output appearance. However, without the guidance of DDIM inversion features, the inversion-free framework is prone to flickering.
Our framework is also inversion-free, but further incorporates intra-frame correspondence, greatly improving temporal consistency while maintaining high controllability.

\section{Methodology}

\subsection{Preliminary}
\label{sec:preliminary}

We follow the inversion-free image translation pipeline of Stable Diffusion based on SDEdit~\cite{meng2021sdedit} and ControlNet~\cite{zhang2023adding}, and adapt it to video translation. 
An input frame $I$ is first mapped to a latent feature $x_0=\mathcal{E}(I)$ with an Encoder $\mathcal{E}$. Then, SDEdit applies DDPM forward process~\cite{ho2020denoising} to add Gaussian noise to $x_0$
\begin{equation}\label{eq:forward_sample}
  q(x_t|x_{0})=\mathcal{N}(x_t;\sqrt{\bar{\alpha}_t}x_0, (1-\bar{\alpha}_t)\mathbf{I}),
\end{equation}
where $\bar{\alpha}_t$ is a pre-defined hyperparamter at the DDPM step $t$.
Then, in the DDPM backward process~\cite{ho2020denoising}, the Stable Diffusion U-Net $\epsilon_\theta$ predicts the noise of the latent feature to iteratively translate $x'_T=x_T$ to $x'_0$ guided by prompt $c$:
\begin{equation}\label{eq:ddpm}
  x'_{t-1}=\frac{\sqrt{\bar\alpha_{t-1}}\beta_t }{1-\bar\alpha_t}\hat{x}'_0 + \frac{(1-\bar\alpha_{t-1})(\sqrt{\alpha_t}x'_t+\beta_tz_t)}{1-\bar\alpha_t},
\end{equation}
where $\alpha_t$ and $\beta_t=1-\alpha_t$ are pre-defined hyperparamters, $z_t$ is a randomly sampled standard
Guassian noise, and $\hat{x}'_0$ is the predicted $x'_0$ at the denoising step $t$,
\begin{equation}\label{eq:denoise}
  \hat{x}'_0=(x'_t-\sqrt{1-\bar\alpha_t}\epsilon_\theta(x'_t, t, c, e))/\sqrt{\bar\alpha_t},
\end{equation}
and $\epsilon_\theta(x_t, t', c, e)$ is the predicted noise of $x'_t$ based on the step $t$, the text prompt  $c$ and the ControlNet condition $e$. The $e$ can be edges, poses or depth maps extracted from $I$ to provide extra structure or layout information.
Finally, the translated frame $I'=\mathcal{D}(x'_0)$ is obtained with a Decoder $\mathcal{D}$. SDEdit allows users to adjust the transformation degree by setting different initial noise level with $T$, \ie, large $T$ for greater appearance variation between $I'$ and $I$.
For simplicity, we will omit the denoising step $t$ in the following.


\subsection{Overall Framework}

The proposed zero-shot video translation pipeline is illustrated in Fig.~\ref{fig:pipeline}. Given a set of video frames $\mathbf{I}=\{I_i\}^N_{i=1}$, 
we follow Sec.~\ref{sec:preliminary} to perform DDPM forward and backward processes to obtain its transformed $\mathbf{I}'=\{I'_i\}^N_{i=1}$.
Our adaptation focuses on incorporating the spatial and temporal correspondences of $\mathbf{I}$ into the U-Net. More specifically, we define temporal and spatial correspondences of $\mathbf{I}$ as:
\begin{itemize}
    \item \textbf{Temporal correspondence}. This inter-frame correspondence is measured by optical flows between adjacent frames, a pivotal element in keeping temporal consistency. Denoting the optical flow and occlusion mask from $I_i$ to $I_j$ as $w^j_i$ and $M^j_i$ respectively, our objective is to ensure that $I'_i$ and $I'_{i+1}$ share $w^{i+1}_i$ in non-occluded regions.
    \item \textbf{Spatial correspondence}. This intra-frame correspondence is gauged by self-similarity among pixels within a single frame. The aim is for $I'_i$ to share self-similarity as $I_i$, \ie, semantically similar content is transformed into a similar appearance, and vice versa. This preservation of semantics and spatial layout implicitly contributes to improving temporal consistency during translation.
\end{itemize}

Our adaptation focuses on the \textit{input feature} and the \textit{attention module} of the decoder layer within the U-Net, since decoder layers are less noisy than encoder layers, and are more semantically meaningful than the $x_t$ latent space:
\begin{itemize}
\item \textbf{Feature adaptation}.~We propose a novel \METHODNAME-aware feature optimization approach as illustrated in Fig.~\ref{fig:pipeline}. We design a spatial consistency loss $\mathcal{L}_{spat}$ and a temporal consistency loss $\mathcal{L}_{temp}$ to directly optimize the decoder-layer features $\mathbf{f}=\{f_i\}_{i=1}^N$ to strengthen their temporal and spatial coherence with the input frames.
\item \textbf{Attention adaptation}.~We replace self-attentions with \METHODNAME-guided attentions, comprising three components, as shown in Fig.~\ref{fig:pipeline}. Spatial-guided attention first aggregates features based on the self-similarity of the input frame. Then, cross-frame attention is used to aggregate features across all frames. Finally, temporal-guided attention aggregates features along the same optical flow to further reinforce temporal consistency.
\end{itemize}

The proposed feature adaptation directly optimizes the feature towards high spatial and temporal coherence with $\mathbf{I}$.
Meanwhile, our attention adaptation indirectly improves coherence by imposing soft constraints on how and where to attend to valid features. We find that combining these two forms of adaptation achieves the best performance.

\subsection{\textbf{\METHODNAME}-Aware Feature Optimization}
\label{sec:feature}

The input feature $\mathbf{f}=\{f_i\}_{i=1}^N$ of each decoder layer of U-Net
is updated by gradient descent through optimizing
\begin{equation}
  \hat{\mathbf{f}}=\arg\min_{\mathbf{f}} \mathcal{L}_{temp}(\mathbf{f}) + \mathcal{L}_{spat}(\mathbf{f}).
\end{equation}
The updated $\hat{\mathbf{f}}$ replaces $\mathbf{f}$ for subsequent processing.

For the temporal consistency loss $\mathcal{L}_{temp}$, we would like the feature values of the corresponding positions between every two adjacent frames to be consistent,
\begin{equation}
  \mathcal{L}_{temp}(\mathbf{f}) = \sum_i\|M_i^{i+1}(f_{i+1}-w_i^{i+1}(f_i))\|_1
\end{equation}

For the spatial consistency loss $\mathcal{L}_{spat}$, we use the cosine similarity in the feature space to measure the spatial correspondence of $I_i$. Specifically, we perform a single-step DDPM forward and backward process over $I_i$, and extract the U-Net decoder feature denoted as $f^r_i$. Since a single-step forward process adds negligible noises, $f^r_i$ can serve as a semantic meaningful representation of $I_i$ to calculate the semantic similarity. Then, the cosine similarity between all pairs of elements can be simply calculated as the gram matrix of the normalized feature. Let $\tilde{f}$ denote the normalized $f$ so that each element of $\tilde{f}$ is a unit vector.
We would like the gram matrix of $\tilde{f}_i$ to approach the gram matrix of $\tilde{f}^r_i$,
\begin{equation}\label{eq:content_loss}
  \mathcal{L}_{spat}(\mathbf{f})= \lambda_\text{spat}\sum_i\|\tilde{f}_i\tilde{f}_i^{~\top}-\tilde{f}^r_i\tilde{f}_i^{r\top}\|^2_2.
\end{equation}

\begin{figure}[t]
\centering
\includegraphics[width=\linewidth]{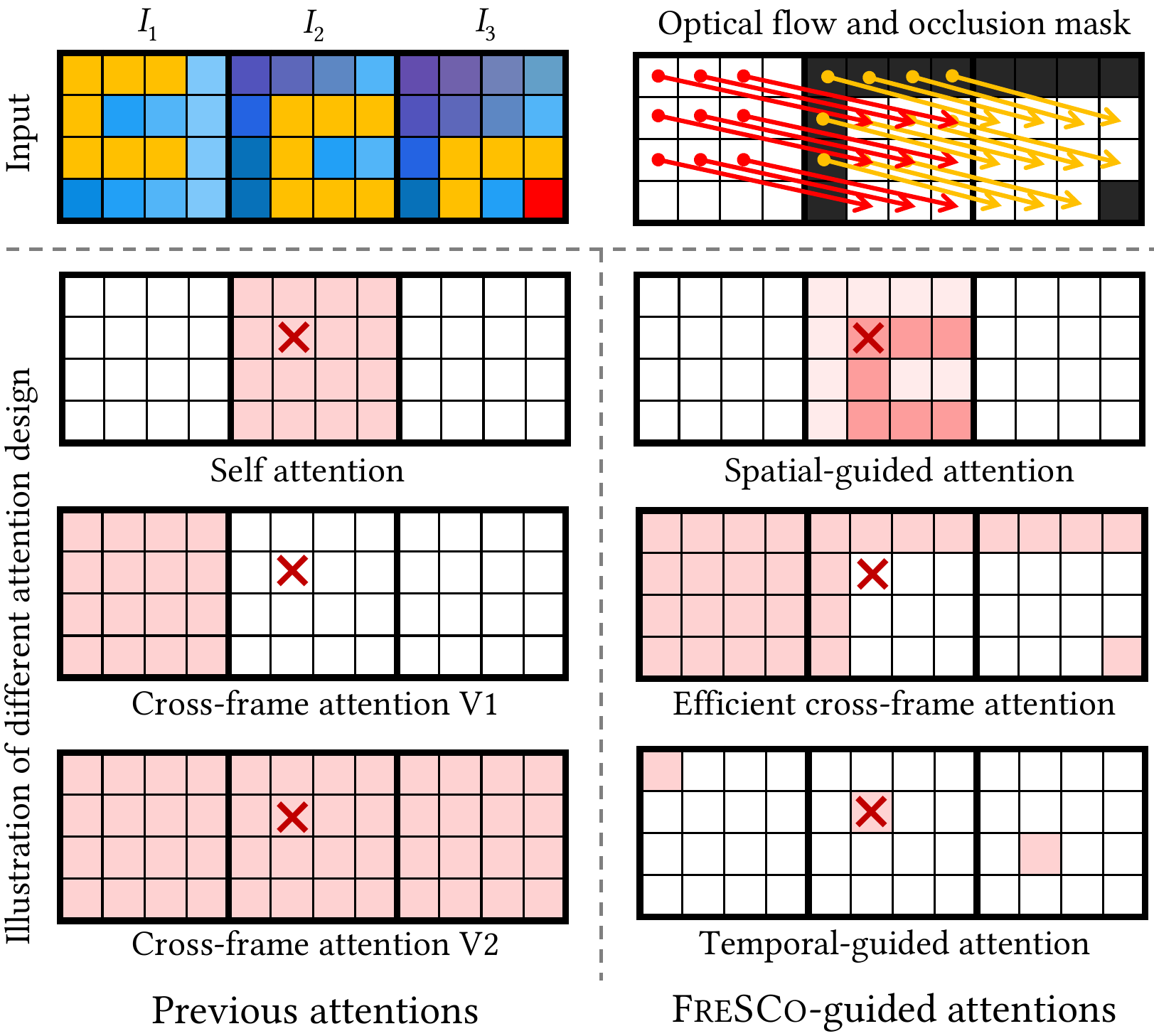}\vspace{-2mm}
\caption{Illustration of attention mechanism. The patches marked with red crosses attend to the colored patches and aggregate their features. Compared to previous attentions, \METHODNAME-guided attention further considers intra-frame and inter-frame correspondences of the input. Spatial-guided attention aggregates intra-frame features based on the self-similarity of the input frame (darker indicates higher weights). Efficient cross-frame attention eliminates redundant patches and retains unique patches. Temporal-guided attention aggregates inter-frame features on the same flow.}\vspace{-4mm}
\label{fig:attention}
\end{figure}

\subsection{\textbf{\METHODNAME}-Guided Attention}
\label{sec:attention}

A \METHODNAME-guided attention layer contains three consecutive modules: spatial-guided attention, efficient cross-frame attention and temporal-guided attention, as shown in Fig.~\ref{fig:pipeline}.

\noindent\textbf{Spatial-guided attention.}~In contrast to self-attention, patches in spatial-guided attention aggregate each other based on the similarity of patches before translation rather than their own similarity. Specifically, consistent with calculating $\mathcal{L}_{spat}$ in Sec.~\ref{sec:feature}, we perform a single-step DDPM forward and backward process over $I_i$, and extract its self-attention query vector $Q^r_i$ and key vector $K^r_i$. 
Then, spatial-guided attention aggregate $Q_i$ with
\begin{equation}
Q'_i=\textit{Softmax}(\frac{Q^r_iK_i^{r\top}}{\lambda_s\sqrt{d}})\cdot Q_i,
\end{equation}
where $\lambda_s$ is a scale factor and $d$ is the query vector dimension. As shown in Fig.~\ref{fig:attention}, the foreground patch will mainly aggregate features in the C-shaped foreground region, and attend less to the background region. As a result, $Q'$ has better spatial consistency with the input frame than $Q$.

\noindent\textbf{Efficient cross-frame attention.}~We replace self-attention with cross-frame attention to regularize the global style consistency.
Rather than using the first frame or the previous frame as reference~\cite{khachatryan2023text2video,ceylan2023pix2video} (V1, Fig.~\ref{fig:attention}), which cannot handle the newly emerged objects (\eg, fingers in Fig.~\ref{fig:challenge}(b)), or using all available frames as reference (V2, Fig.~\ref{fig:attention}), which is computationally inefficient, we aim to consider all frames simultaneously but with as little redundancy as possible. Thus, we propose efficient cross-frame attentions: Except for the first frame, we only reference to the areas of each frame that were not seen in its previous frame (\ie, the occlusion region). Thus, we can construct a cross-frame index $p_u$ of all patches within the above region. Keys and values of these patches can be sampled as $K[p_u]$, $V[p_u]$. Then, cross-frame attention is applied
\begin{equation}
V'_i=\textit{Softmax}(\frac{Q'_i(K[p_u])^\top}{\sqrt{d}})\cdot V[p_u].
\end{equation}

\noindent\textbf{Temporal-guided attention.}~Inspired by FLATTEN~\cite{cong2023flatten}, we use flow-based attention to regularize fine-level cross-frame consistency. We trace the same patches in different frames as in Fig.~\ref{fig:attention}. For each optical flow, we build a cross-frame index $p_f$ of all patches on this flow. In FLATTEN, each patch can only attend to patches in other frames, which is unstable when a flow contains few patches. Different from it, the temporal-guided attention has no such limit,
\begin{equation}
H[p_f]=\textit{Softmax}(\frac{Q[p_f](K[p_f])^\top}{\lambda_t\sqrt{d}})\cdot V'[p_f],
\end{equation}
where $\lambda_t$ is a scale factor. And $H$ is the final output of our \METHODNAME-guided attention layer.

\subsection{Long Video Translation}

The number of frames $N$ that can be processed at one time is limited by GPU memory. For long video translation, we follow Rerender-A-Video~\cite{yang2023rerender} to perform zero-shot video translation on keyframes only and use Ebsynth~\cite{jamrivska2019stylizing} to interpolate non-keyframes based on translated keyframes.

\noindent\textbf{Keyframe selection.} Rerender-A-Video~\cite{yang2023rerender} uniformly samples keyframes, which is suboptimal. We propose a heuristic keyframe selection algorithm as summized in Algorithm~\ref{alg:algorithm1}. We relax the fixed sampling step to an interval $[s_\text{min}, s_\text{max}]$, and densely sample keyframes when motions are large (measured by $L_2$ distance between frames).

\setlength{\textfloatsep}{8pt}
\begin{algorithm}[t]
  \caption{Keyframe selection}
  \textbf{Input:} Video $\mathbf{I}=\{I_i\}^M_{i=1}$, sample parameters $s_\text{min}$, $s_\text{max}$\\
  \textbf{Output:} Keyframe index list $\Omega$ in ascending order
  \begin{algorithmic}[1]
    \State initialize $\Omega=[1, M]$ and $d_i=0, \forall i\in[1,M]$
    \State set $d_i=L_2(I_i,I_{i-1}), \forall i\in[s_\text{min}+1,N-s_\text{min}]$
    \While {\textbf{exists} $i$ \textbf{such that} $\Omega[i+1]-\Omega[i]>s_\text{max}$}
        \State $\Omega.$\texttt{insert}$(\hat{i}).$\texttt{sort}$()$ with $\hat{i}=\arg\max_i(d_i)$
        \State set $d_j=0$, $\forall~j\in(\hat{i}-s_\text{min}, \hat{i}+s_\text{min})$
    \EndWhile
  \end{algorithmic}
  \label{alg:algorithm1}
\end{algorithm}

\noindent\textbf{Keyframe translation.} With over $N$ keyframes, we split them into several $N$-frame batches. Each batch includes the first and last frames in the previous batch to impose inter-batch consistency, \ie, keyframe indexes of the $k$-th batch are
$\{1,(k-1)(N-2)+2,(k-1)(N-2)+3,...,k(N-2)+2\}$.
Besides, throughout the whole denoising steps, we record the latent features $x'_t$ (Eq.~(\ref{eq:ddpm})) of the first and last frames of each batch, and use them to replace the corresponding latent features in the next batch.

\begin{figure*}[htbp]
\centering
\includegraphics[width=0.98\linewidth]{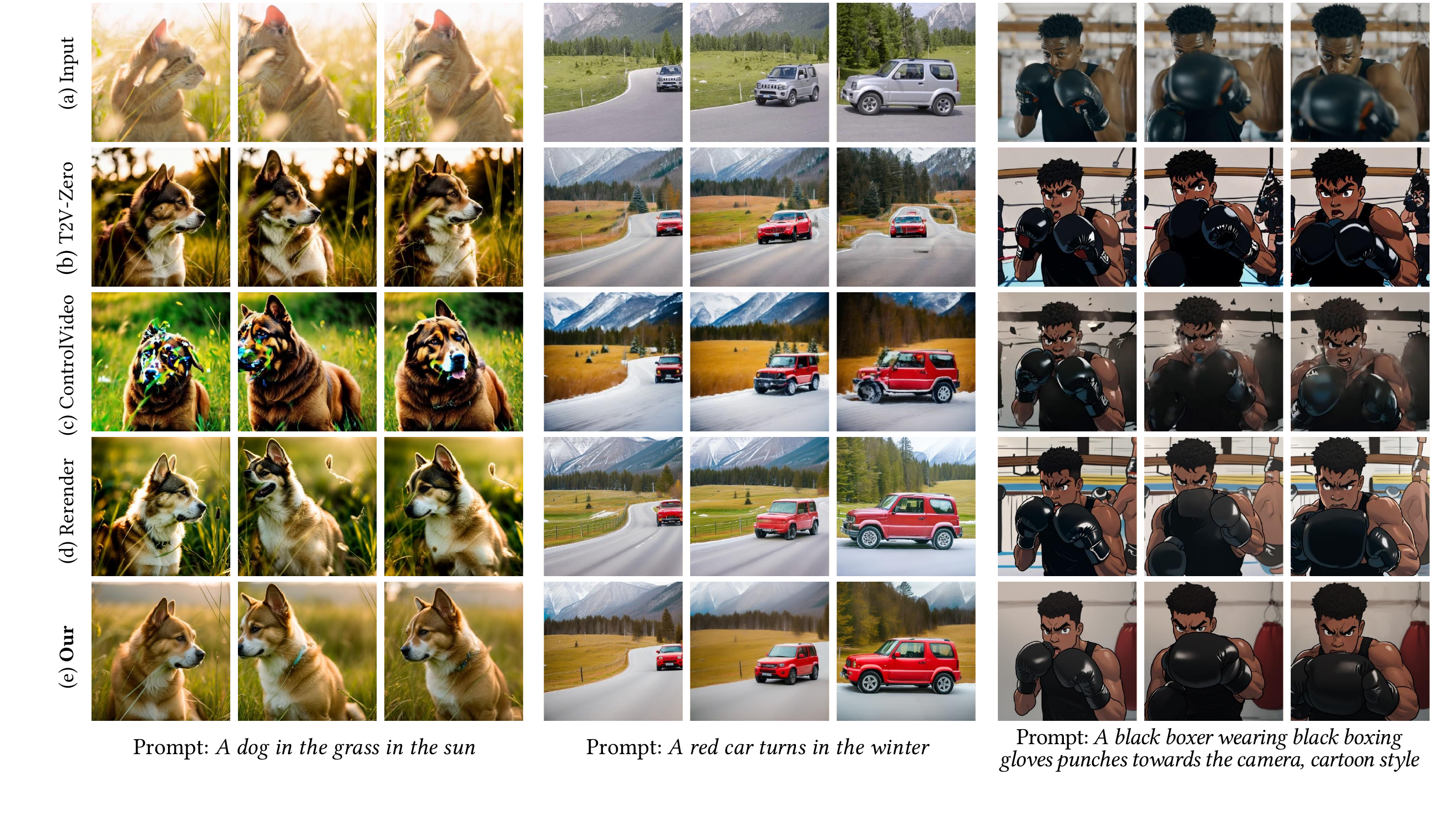}\vspace{-2mm}
\caption{Visual comparison with inversion-free zero-shot video translation methods.}\vspace{-3mm}
\label{fig:compare}
\end{figure*}

\section{Experiments}


\textbf{Implementation details}. The experiment is conducted on one NVIDIA Tesla V100 GPU.
By default, we set batch size $N\in[6,8]$ based on the input video resolution, the loss weight $\lambda_\text{spat}=50$, the scale factors $\lambda_s=\lambda_t=5$. For feature optimization, we update $\mathbf{f}$ for $K=20$ iterations with Adam optimizer and learning rate of $0.4$. \CR{We find optimization mostly converges when $K=20$ and larger $K$ does not bring obvious gains.}
GMFlow~\cite{xu2022gmflow} is used to estimate optical flows and occlusion masks. Background smoothing~\cite{khachatryan2023text2video} is applied to improve temporal consistency in the background region. 

\subsection{Comparison with State-of-the-Art Methods}

We compare with three recent inversion-free zero-shot methods: Text2Video-Zero~\cite{khachatryan2023text2video}, ControlVideo~\cite{zhang2023controlvideo}, Rerender-A-Video~\cite{yang2023rerender}.
To ensure a fair comparison, all methods employ identical settings of ControlNet, SDEdit, and LoRA.
As shown in Fig.~\ref{fig:compare}, all methods successfully translate videos according to the provided text prompts.
However, the inversion-free methods, relying on ControlNet conditions, may experience a decline in video editing quality if the conditions are of low quality, due to issues like defocus or motion blur. For instance, ControlVideo fails to generate a plausible appearance of the dog and the boxer. Text2Video-Zero and Rerender-A-Video struggle to maintain the cat's pose and the structure of the boxer's gloves. In contrast, our method can generate consistent videos based on the proposed
robust \METHODNAME~guidance.

For quantitative evaluation, adhering to standard practices~\cite{qi2023fatezero, ceylan2023pix2video, yang2023rerender}, we employ the evaluation metrics of Fram-Acc (CLIP-based frame-wise editing accuracy),
Tmp-Con (CLIP-based cosine similarity between consecutive frames) and Pixel-MSE (averaged mean-squared pixel error between aligned
consecutive frames).
\CR{We further report Spat-Con ($L_{spat}$ on VGG features) for spatial coherency.}
The results averaged across \CR{23} videos are reported in Table~\ref{tb:quantitative_evaluation}. Notably, our method attains the best editing accuracy and temporal consistency.
We further conduct a user study with \CR{57} participants. Participants are tasked with selecting the most preferable results among the four methods. Table~\ref{tb:quantitative_evaluation} presents the average preference rates across the 11 test videos, revealing that our method emerges as the most favored choice.

\begin{table} []
\caption{\CR{Quantitative comparison and user preference rates.}}\vspace{-2mm}
\label{tb:quantitative_evaluation}
\resizebox{\linewidth}{!}{
\centering
\begin{tabular}{l|c|c|c|c|c}
\toprule
Metric & Fram-Acc $\uparrow$ & Tem-Con $\uparrow$ & Pixel-MSE $\downarrow$ & Spat-Con $\downarrow$  & User $\uparrow$  \\
\midrule
T2V-Zero & 0.918 & 0.965 & 0.038 & 0.0845 & 9.1\% \\
ControlVideo & 0.932 & 0.951 & 0.066 & 0.0957 & 2.6\% \\
Rerender & 0.955 & 0.969 & 0.016 & 0.0836 & 23.3\% \\
Ours & \textbf{0.978} & \textbf{0.975} & \textbf{0.012} & \textbf{0.0805} & \textbf{65.0\%}\\
\bottomrule
\end{tabular}}\vspace{-2mm}
\end{table}

\subsection{Ablation Study}

\begin{figure}[t]
\centering
\includegraphics[width=\linewidth]{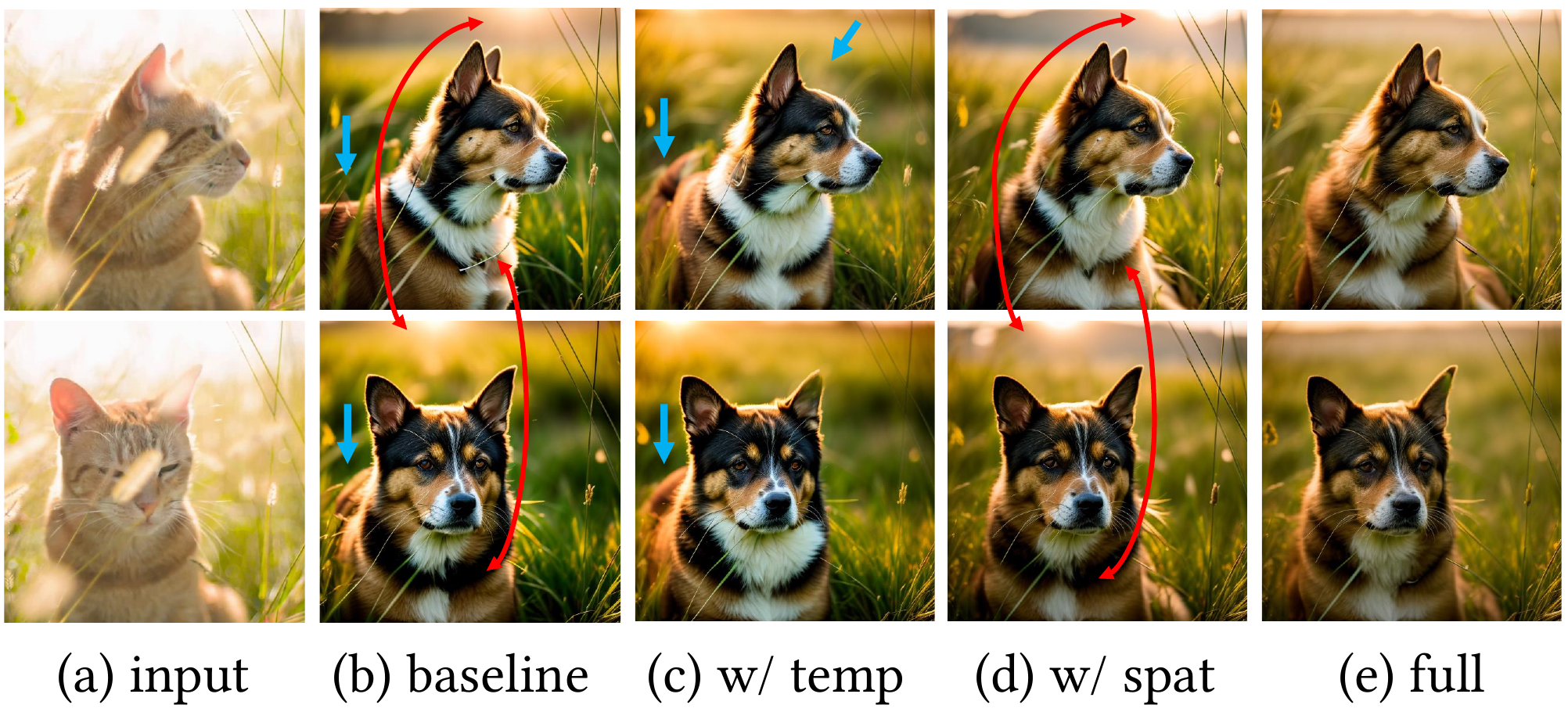}\vspace{-2.5mm}
\caption{Effect of incorporating spatial and temporal correspondences. The blue arrows indicate the spatial inconsistency with the input frames. The red arrows indicate the temporal inconsistency between two output frames.}\vspace{-2.5mm}
\label{fig:ablation1}
\end{figure}

To validate the contributions of different modules to the overall performance, we systematically deactivate
specific modules in our framework. Figure~\ref{fig:ablation1} illustrates the effect of incorporating spatial and temporal
correspondences. The baseline method solely uses cross-frame attention for temporal consistency.
By introducing the temporal-related adaptation, we observe improvements in consistency, such as the alignment of textures and the stabilization of the sun's position across two frames.
Meanwhile, the spatial-related adaptation aids in preserving the pose during translation.

\begin{table}[t]
\begin{center}
\caption{Quantitative ablation study.}\vspace{-2mm}
\resizebox{\linewidth}{!}{
\begin{tabular}{l|cccccc}
\toprule
\textbf{Metric} & baseline & w/ temp & w/ spat & w/ attn & w/ opt & full\\
\midrule
Fram-Acc $\uparrow$ & \textbf{1.000} & \textbf{1.000} & \textbf{1.000} & \textbf{1.000} & \textbf{1.000} & \textbf{1.000} \\
Tem-Con $\uparrow$ & 0.974 & 0.979 & 0.976 & 0.976 & 0.977 & \textbf{0.980} \\
Pixel-MSE $\downarrow$ & 0.032 & 0.015 & 0.020 & 0.016 & 0.019 & \textbf{0.012} \\
\bottomrule
\end{tabular}}\vspace{-4mm}
\label{tb:quantitative_ablation}
\end{center}
\end{table}

\begin{figure}[tbp]
\centering
\includegraphics[width=\linewidth]{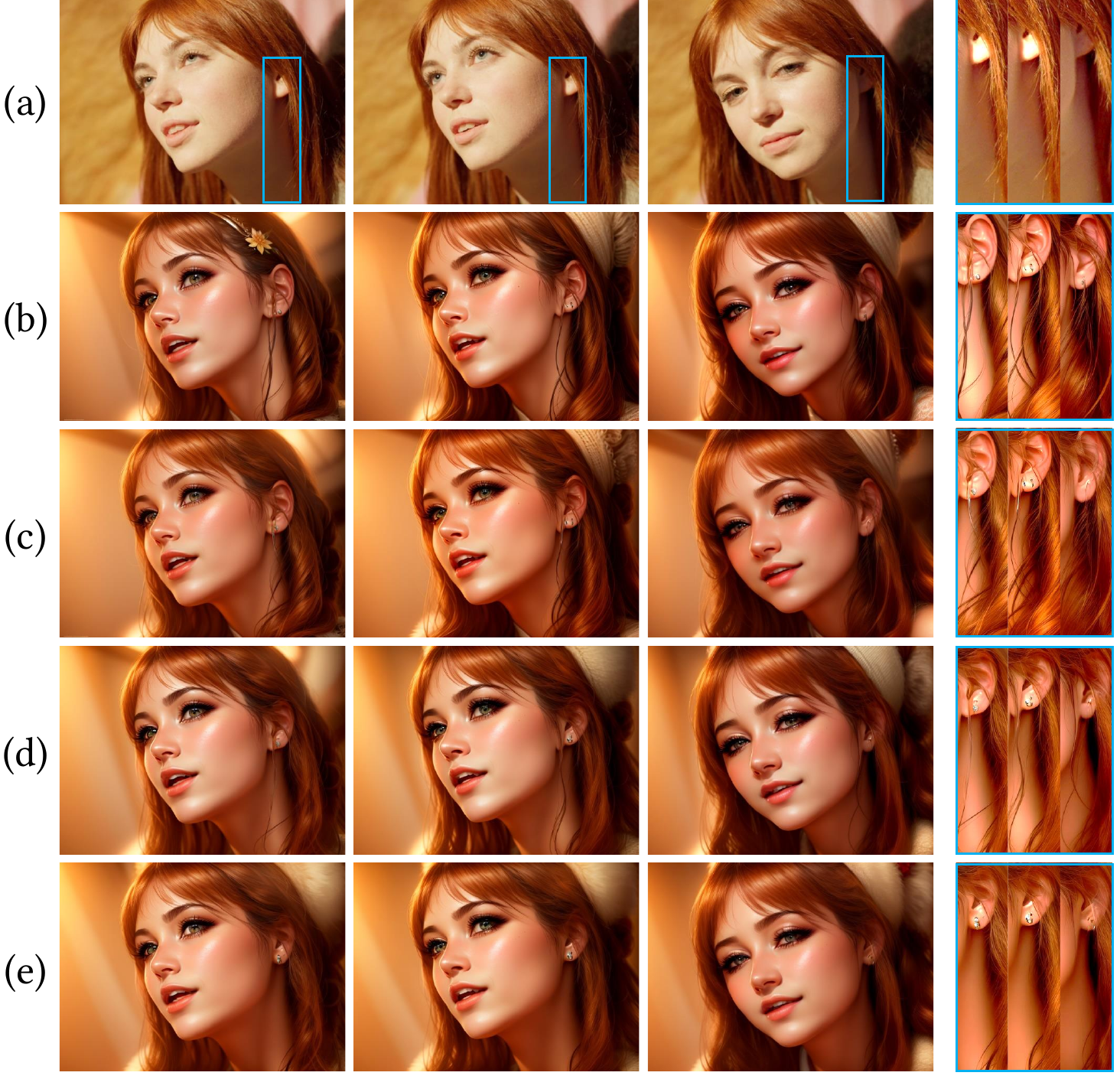}\vspace{-2.5mm}
\caption{Effect of attention adaptation and feature adaptation. Top row: (a) Input. Other rows: Results obtained with (b) only cross-frame attention, (c) attention adaptation, (d) feature adaptation, (e) both attention and feature adaptations, respectively. The blue region is enlarged with its contrast enhanced on the right for better comparison. Prompt: A beautiful woman in CG style.}\vspace{-1mm}
\label{fig:ablation_opt}
\end{figure}

\begin{figure}[tbp]
\centering
\includegraphics[width=\linewidth]{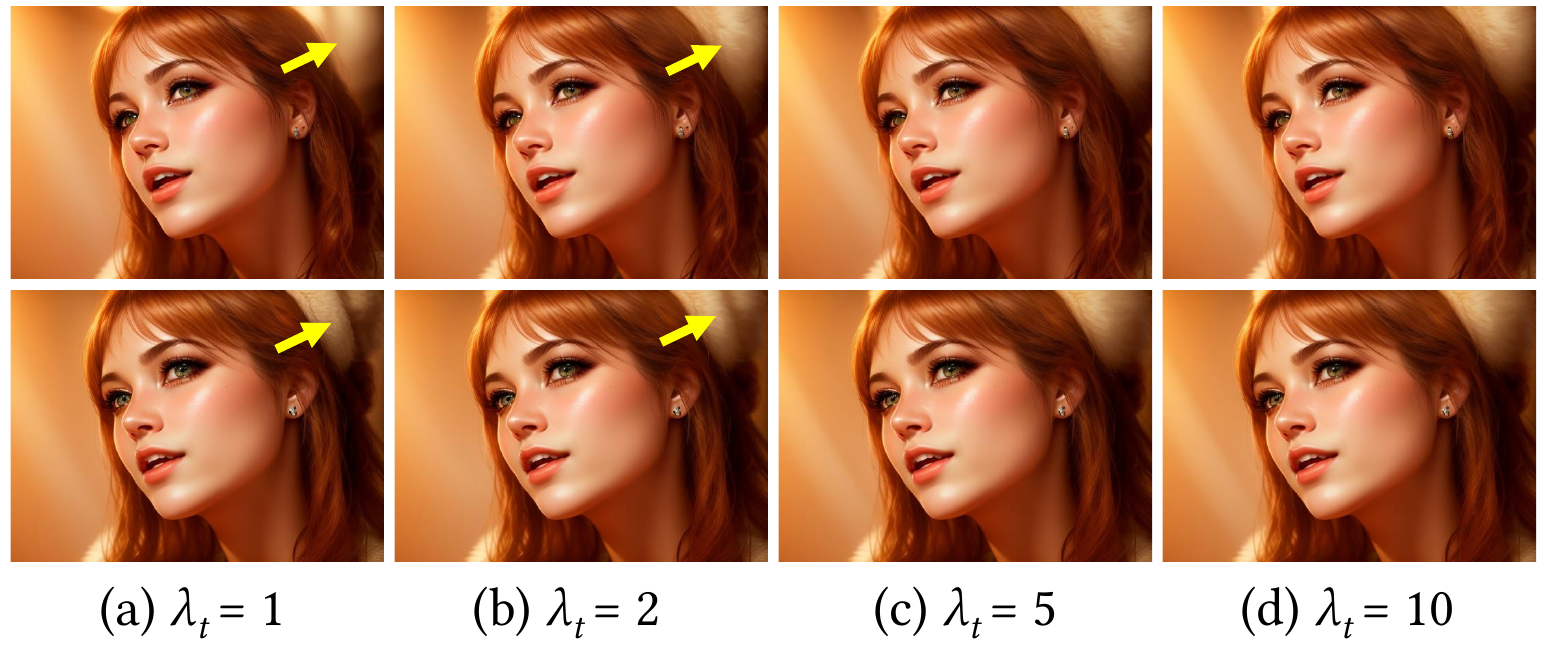}\vspace{-2.5mm}
\caption{Effect of $\lambda_t$. \CR{Quantitatively, the Pixel-MSE scores are (a) 0.016, (b) 0.014, (c) 0.013, (d) 0.012.}
The yellow arrows indicate the inconsistency between the two frames.}\vspace{-3mm}
\label{fig:ablation_lambdat}
\end{figure}

\begin{figure}[tbp]
\centering
\includegraphics[width=\linewidth]{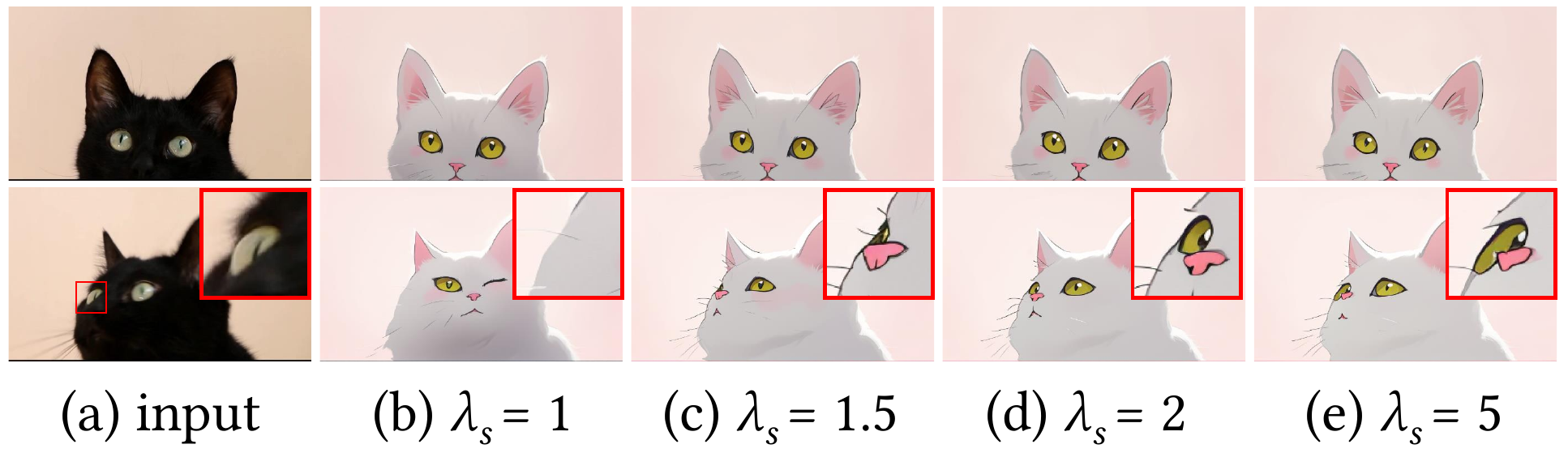}\vspace{-2.5mm}
\caption{Effect of $\lambda_s$. The region in the red box is enlarged and shown in the top right for better comparison. Prompt: A cartoon white cat in pink background.}\vspace{-3mm}
\label{fig:ablation_lambdas}
\end{figure}

\begin{figure}[tbp]
\centering
\includegraphics[width=\linewidth]{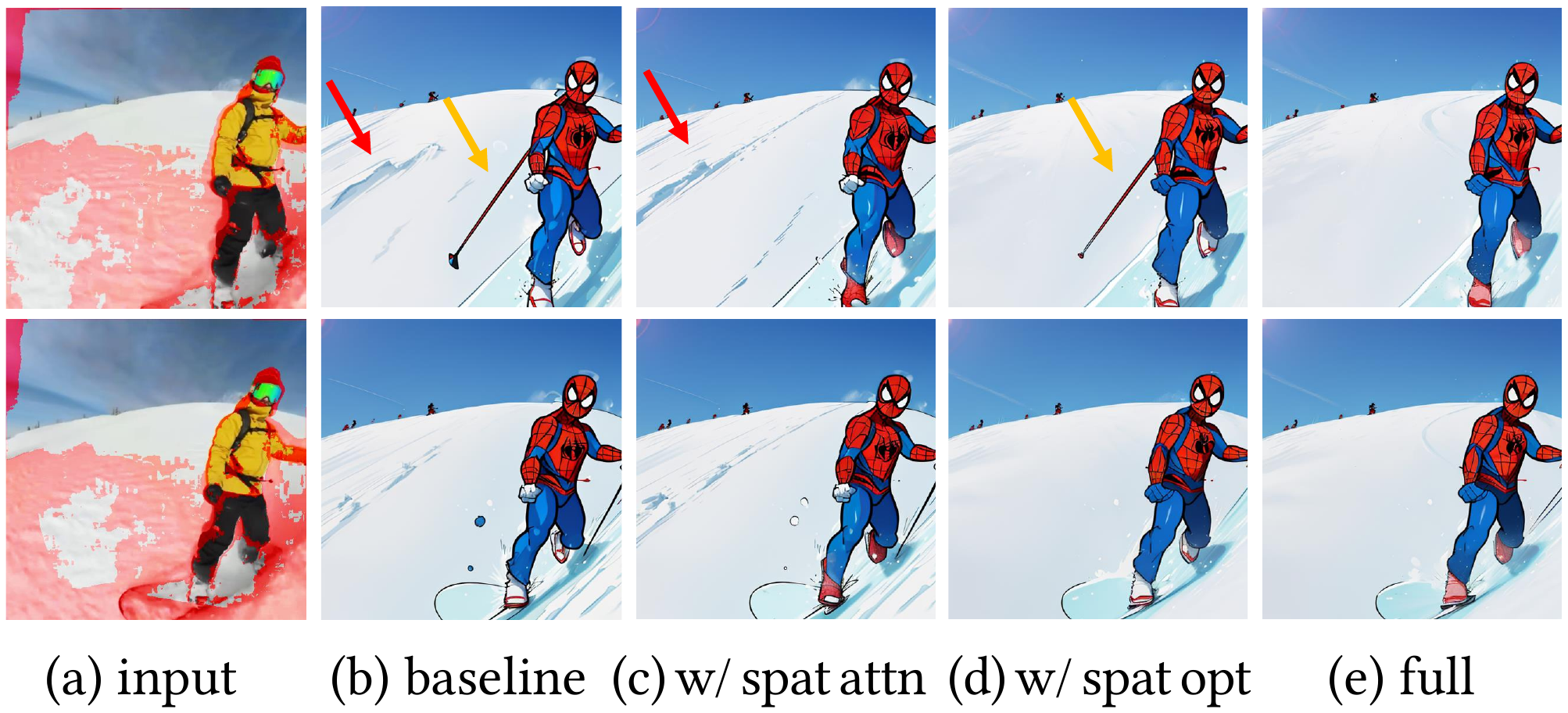}\vspace{-2.5mm}
\caption{Effect of incorporating spatial correspondence. (a) Input covered with red occlusion mask. (b)-(d) Our spatial-guided attention and spatial consistency loss help reduce the inconsistency in ski poles (yellow arrows) and snow textures (red arrows), respectively. Prompt: A cartoon Spiderman is skiing.}\vspace{-2mm}
\label{fig:ablation_spat}
\end{figure}

In Fig.~\ref{fig:ablation_opt}, we study the effect of attention adaptation and feature adaption. Clearly, each enhancement individually improves temporal consistency to a certain extent, but neither achieves perfection. Only the combination of the two completely eliminates the inconsistency observed in hair strands\CR{, which is quantitatively verified by the Pixel-MSE scores of 0.037, 0.021, 0.018, 0.015 for Fig.~\ref{fig:ablation_opt}(b)-(e), respectively.}
Regarding attention adaptation, we further delve into temporal-guided attention and spatial-guided attention. The strength of the constraints they impose is determined by $\lambda_t$ and $\lambda_s$, respectively. As shown in Figs.~\ref{fig:ablation_lambdat}-\ref{fig:ablation_lambdas}, an increase in $\lambda_t$  effectively enhances consistency between two transformed frames in the background region, while an increase in $\lambda_s$ boosts pose consistency between the transformed cat and the original cat.
Beyond spatial-guided attention, our spatial consistency loss also plays an important role, as validated in Fig.~\ref{fig:ablation_spat}. In this example, rapid motion and blur make optical flow hard to predict, leading to a large occlusion region. Spatial correspondence guidance is particularly crucial to constrain the rendering in this region.
Clearly, each adaptation makes a distinct contribution, such as eliminating the unwanted ski pole and inconsistent snow textures. Combining the two yields the most coherent results\CR{, as quantitatively verified by the Pixel-MSE scores of 0.031, 0.028, 0.025, 0.024 for Fig.~\ref{fig:ablation_spat}(b)-(e), respectively.}


Table~\ref{tb:quantitative_ablation} provides a quantitative evaluation of the impact of each module.
In alignment with the visual results, it is evident that each module contributes to the overall enhancement of temporal consistency. Notably, the combination of all adaptations yields the best performance.

\begin{figure}[tbp]
\centering
\includegraphics[width=\linewidth]{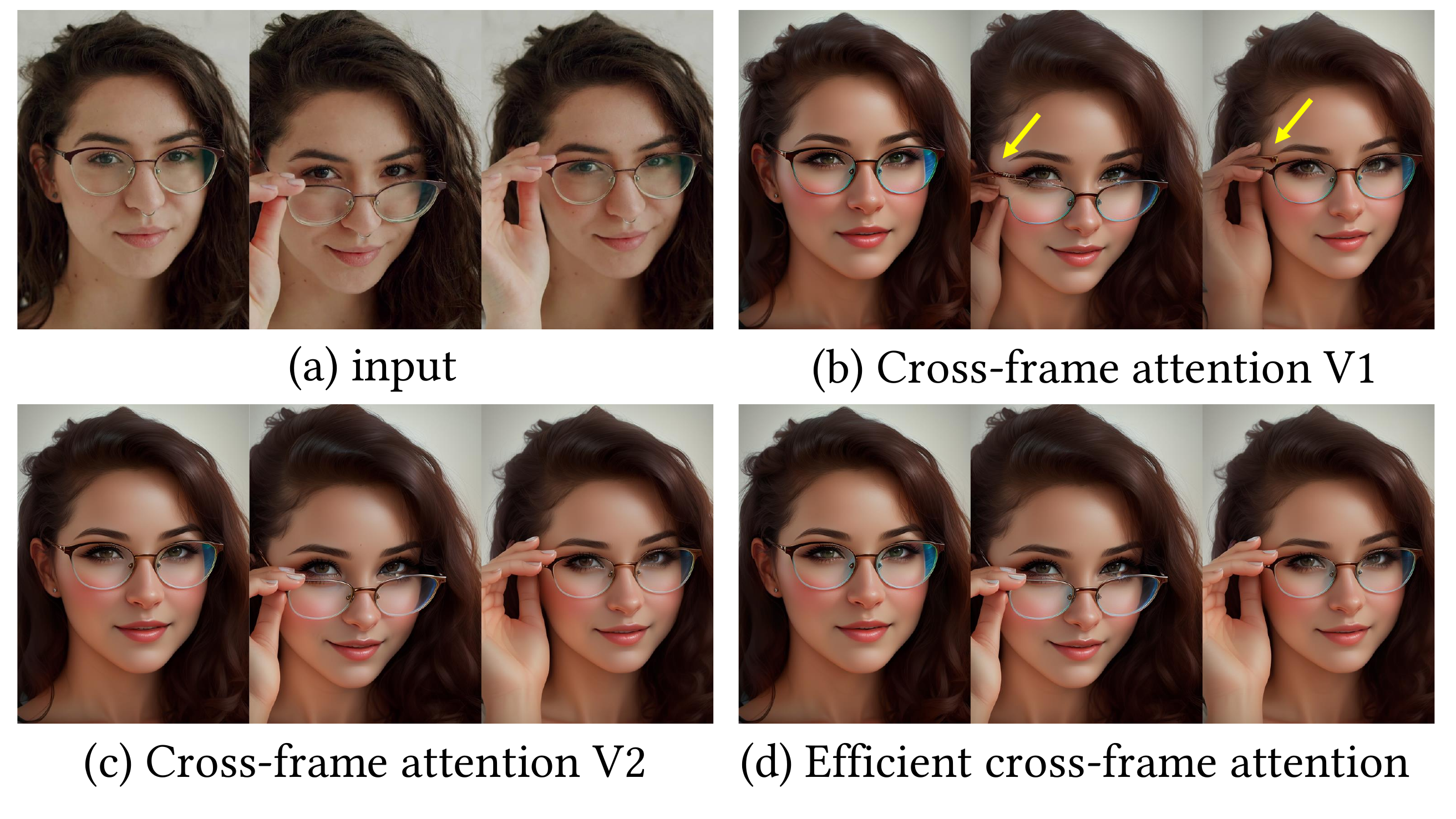}\vspace{-2.5mm}
\caption{Effect of efficient cross-frame attention. (a) Input. (b) Cross-frame attention V1 attends to the previous frame only, thus failing to synthesize the newly appearing fingers. (d) The efficient cross-frame attention achieves the same performance as (c) cross-frame attention V2, but reduces the region that needs to be attended to by $41.6\%$ in this example. 
Prompt: A beautiful woman holding her glasses in CG style.}\vspace{-2mm}
\label{fig:ablation_CF}
\end{figure}

\begin{figure}[tbp]
\centering
\includegraphics[width=\linewidth]{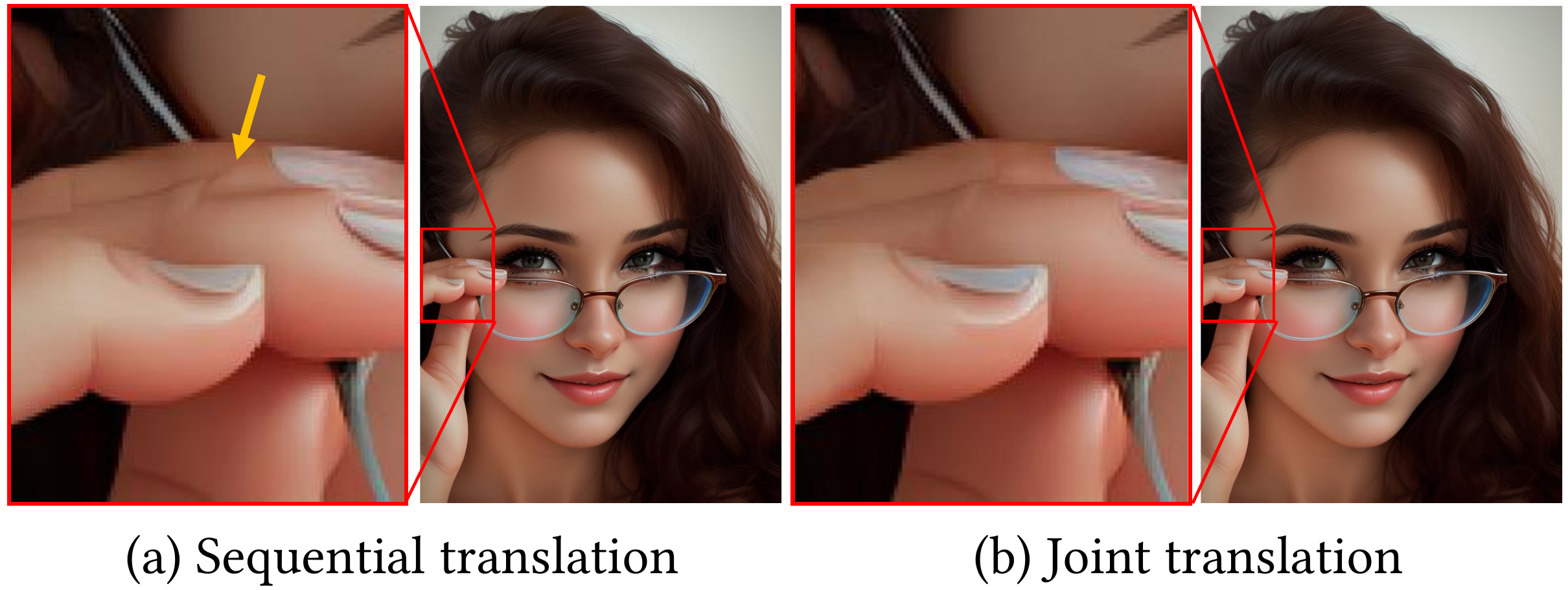}\vspace{-2.5mm}
\caption{Effect of joint multi-frame translation. Sequential translation relies on the previous frame alone. Joint translation uses all frames in a batch to guide each other, thus achieving accurate finger structures by referencing to the third frame in Fig.~\ref{fig:ablation_CF}}\vspace{-2mm}
\label{fig:ablation_joint}
\end{figure}

Figure~\ref{fig:ablation_CF} ablates the proposed efficient cross-frame attention. As with Rerender-A-Video in Fig.~\ref{fig:challenge}(b), sequential frame-by-frame translation is vulnerable to new appearing objects. Our cross-frame attention allows attention to all unique objects within the batched frames, which is not only efficient but also more robust, as demonstrated in Fig.~\ref{fig:ablation_joint}.

\begin{figure}[tbp]
\centering
\includegraphics[width=\linewidth]{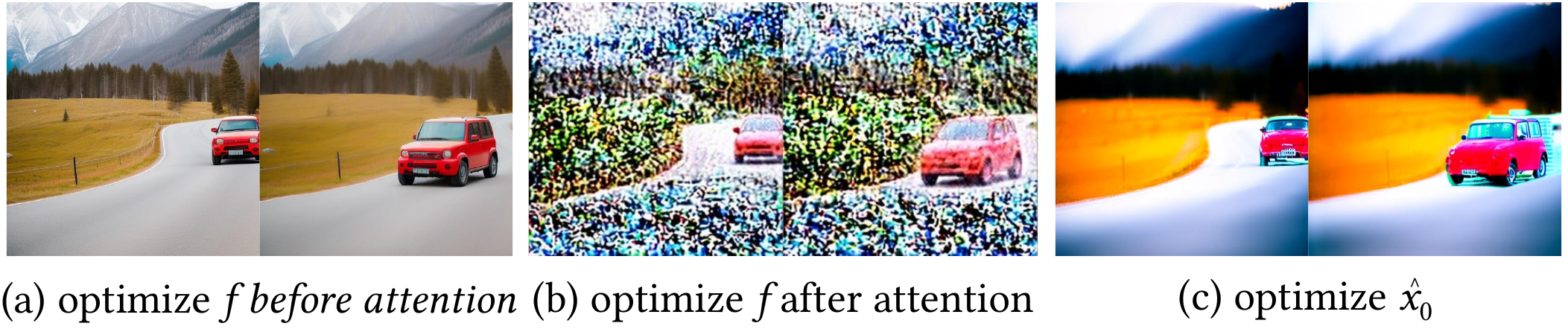}\vspace{-2mm}
\caption{Diffusion features to optimize.}\vspace{-2mm}
\label{fig:ablation_optimize}
\end{figure}

\CR{\METHODNAME~uses diffusion features before the attention layers for optimization. Since U-Net is trained to predict noise, features after attention layers (near output layer) are noisy, leading to failure optimization (Fig.~\ref{fig:ablation_optimize}(b)). Meanwhile, the four-channel $\hat{x}'_0$ (Eq.~(\ref{eq:denoise})) is highly compact, which is not suitable for warping or interpolation. Optimizing $\hat{x}'_0$ results in severe blurs and over-saturation artifacts (Fig.~\ref{fig:ablation_optimize}(c)).}

\subsection{More Results}

\begin{figure}[tbp]
\centering
\includegraphics[width=\linewidth]{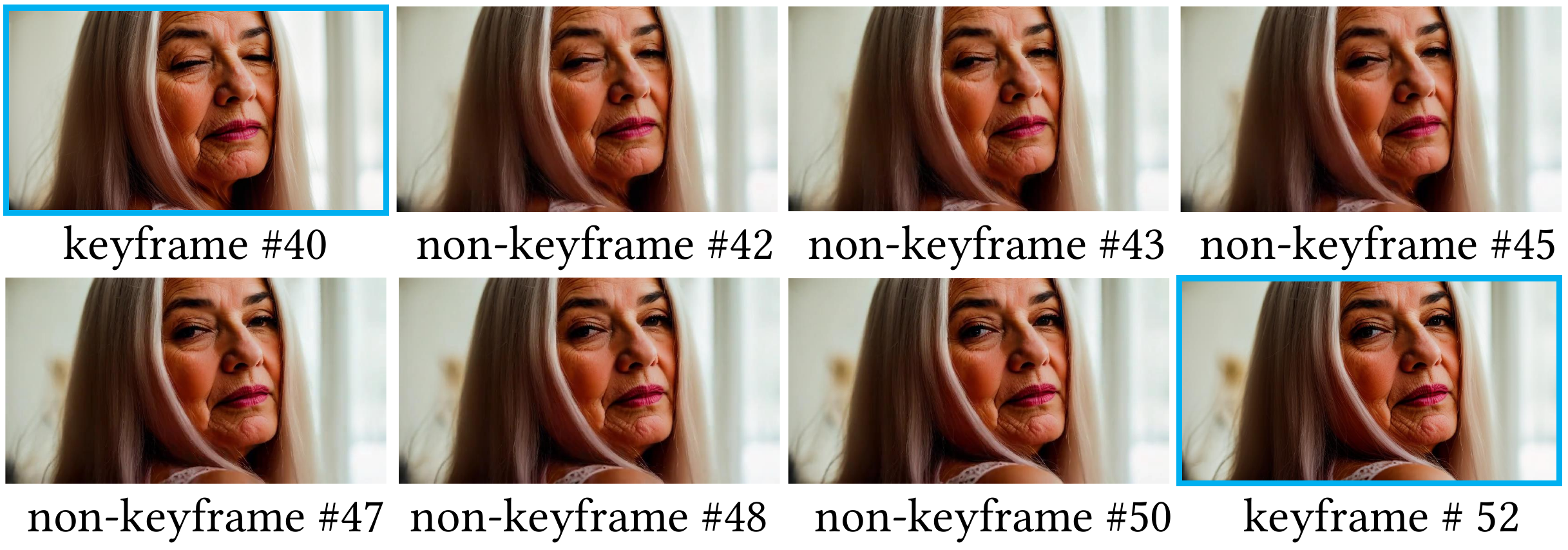}\vspace{-2.5mm}
\caption{Long video generation by interpolating non-keyframes based on the translated keyframes.}\vspace{-3mm}
\label{fig:long_video}
\end{figure}

\noindent\textbf{Long video translation.}~Figure~\ref{fig:teaser} presents examples of long video translation.~A 16-second video comprising $400$ frames are processed, where $32$ frames are selected as keyframes for diffusion-based translation and the remaining  $368$ non-keyframes are interpolated. Thank to our \METHODNAME~guidance to generate coherent keyframes, the non-keyframes exhibit coherent interpolation as in
Fig.~\ref{fig:long_video}.

\begin{figure}[tbp]
\centering
\includegraphics[width=\linewidth]{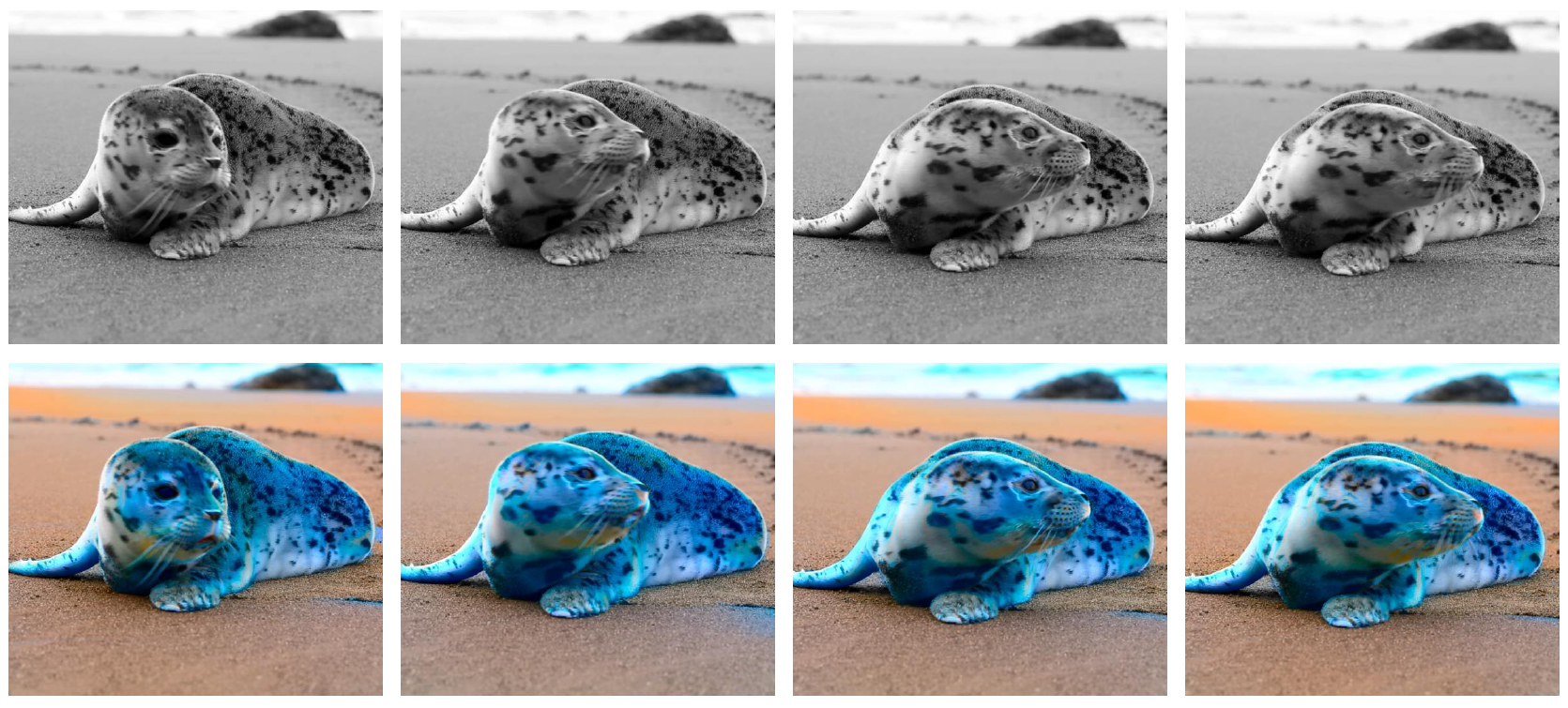}\vspace{-2.5mm}
\caption{Video colorization. Prompt: A blue seal on the beach.}\vspace{-1mm}
\label{fig:color}
\end{figure}

\noindent\textbf{Video colorization.} Our method can be applied to video colorization. As shown in Fig.~\ref{fig:color}, by combining the L channel from the input and the AB channel from the translated video, we can colorize the input without altering its content.


\subsection{Limitation and Future Work}\vspace{-1.5mm}

%
\CR{In terms of limitations,}
first, Rerender-A-Video directly aligns frames at the pixel level, which outperforms our method given high-quality optical flow. We would like to explore an adaptive combination of these two methods in the future to harness the advantages of each.
Second, by enforcing spatial correspondence consistency with the input video, our method does not support large shape deformations and significant appearance changes.
Large deformation makes it challenging to use the optical flow of the original video as a reliable prior for natural motion. This limitation is inherent in zero-shot models. A potential future direction is to incorporate learned motion priors~\cite{guo2023animatediff}.
%

\section{Conclusion}\vspace{-0.5mm}
This paper presents a zero-shot framework to adapt image diffusion models for video translation.~We demonstrate the vital role of preserving intra-frame spatial correspondence, in conjunction with inter-frame temporal correspondence, which is less explored in prior zero-shot methods. Our comprehensive experiments validate the effectiveness of our method in translating high-quality and coherent videos. The proposed \METHODNAME~constraint exhibits
high compatibility with existing image diffusion techniques, suggesting its potential application in other text-guided video editing tasks, such as video super-resolution and colorization.\vspace{1mm}

\small{
\noindent
\textbf{Acknowledgments.} This study is supported under the RIE2020 Industry Alignment Fund Industry Collaboration Projects (IAF-ICP) Funding Initiative, as well as cash and in-kind contribution from the industry partner(s).
This study is also supported by NTU NAP and MOE AcRF Tier 2 (T2EP20221-0012).
}

{
\small
\bibliographystyle{ieeenat_fullname}
\bibliography{sample-bibliography}
}


\end{document}